\documentclass[conference,a4paper]{APSIPA2021}
\usepackage{amsmath}
\usepackage{graphicx}
\usepackage{multirow}
\usepackage{threeparttable}

\usepackage{geometry}
\usepackage{fancyhdr}

\fancypagestyle{firststyle}{
  \fancyhf{}
  \fancyhead[C]{2024 Asia Pacific Signal and Information Processing Association Annual Summit and Conference (APSIPA ASC)}
}

\begin{document}

\title{Green Video Camouflaged Object Detection} 

\author{
\authorblockN{
Xinyu Wang\authorrefmark{1},
Hong-Shuo Chen\authorrefmark{1},
Zhiruo Zhou\authorrefmark{1},
Suya You\authorrefmark{1},
Azad M. Madni\authorrefmark{1} and
C.-C. Jay Kuo\authorrefmark{1}
}

\authorblockA{
\authorrefmark{1}
University of Southern California, USA \\
\{xwang350, hongshuo, zhiruozh, suya, azad.madni, jckuo\}@usc.edu}
}

\maketitle
\thispagestyle{firststyle}
\pagestyle{fancy}

\begin{abstract}
  Camouflaged object detection (COD) aims to distinguish hidden objects embedded in an environment highly similar to the object.
  Conventional video-based COD (VCOD) methods explicitly extract motion cues or employ complex deep learning networks to handle the temporal information, which is limited by high complexity and unstable performance. In this work, we propose a green VCOD method named GreenVCOD. Built upon a green ICOD method,  
  GreenVCOD uses long- and short-term temporal neighborhoods (TN) to capture joint spatial/temporal context information for decision refinement. Experimental results show that GreenVCOD offers competitive performance compared to state-of-the-art VCOD benchmarks.
\end{abstract}

\section{Introduction}

Video Camouflaged Object Detection (VCOD) identifies objects that blend into their surroundings within video footage. Unlike image camouflaged object detection, VCOD needs to address more challenges posed by object movement, camera shaking, constantly changing environments, and the inherent variability and complexity of video scenes. In addition, the hidden object has a color or texture pattern highly similar to the background, further increasing the difficulty compared to traditional target detection and tracking tasks. Despite its challenges, VCOD is essential for a wide range of applications, such as surveillance and security \cite{8628238}, medical imaging \cite{fan2020camouflaged, fan2020infnet, fan2020pranet}, and autonomous driving.

Existing video detection approaches have demonstrated the importance of incorporating temporal motion cues of target objects. Traditional techniques, such as frame difference and optical flow, explicitly capture motion information through homo-graphic calculation and angle estimation. These motion patterns suggest that background noise easily influences the position and motion trajectory of the camouflaged object and may lead to the accumulation of errors. Implicit methods utilizing feature or prediction correlations have higher self-correction capabilities, but the learning results are highly dependent on the training samples and lack generalization ability. Also, the implicit motion cues are not explicitly evaluated, which makes the learned motion cues less reliable \cite{zhang2024explicit}. Consequently, extracting and incorporating frame-level spatial information and temporal motion cues improves detection accuracy and efficiency.

Additionally, given the applications in wildlife species protection and security operations, developing compact models for video object detection is essential to facilitate edge computing and enhance flexibility in mobile devices. Current models, which typically employ encoder-decoder architectures or deep neural networks, often require additional motion estimation modules. This requirement can significantly increase the computational cost, highlighting the need for more streamlined solutions that maintain performance while reducing resource demands.

In this work, we propose GreenVCOD, a novel green learning framework to deal with the VCOD problem feed-forwardly. The framework comprises two essential components: the initial-stage prediction module and the temporal refinement module. The problem is first reduced into a series of image-based detection tasks to obtain the initial prediction maps. A cascaded multi-scale decision-boosting network is utilized to refine the prediction graph as the resolution increases gradually. Next, we incorporate temporal information to refine the initial predictions. Instead of explicit motion estimation, our temporal neighborhood prediction cube fixes the focus on a certain spatial location relative to the whole frame. It captures the scene change among adjacent frames, providing implicit spatial-temporal context information without additional computational cost. With features extracted at the initial stage, an additional XGBoost classifier is trained for temporal decision refinement. We build a parallel structure to balance the variety of object movements, including short-term and long-term refinement modules. In the concluding phase, a decision ensemble module is employed to integrate the outcomes derived from both the short-term and long-term refinement modules, thereby ensuring robustness and accuracy in the final result.

To summarize, this work has three significant contributions. First, we propose a novel framework, GreenVCOD, that combines frame-based camouflage detection with temporal refinement, realizing effective video camouflage detection in a feed-forward manner. Second, we propose a temporal neighborhood prediction method. It extracts spatial information and motion cues while requiring no extra computational cost. Third, we build a parallel short - and long-term decision refinement structure that makes a good trade-off between different object motion patterns. 

The rest of this paper is organized as follows. The related work is reviewed in Sec. \ref{sec:Background}. The GreenVCOD method is described in Sec. \ref{sec:method}.  Experimental results are shown in Sec.\ref{sec:experiment}. Finally, the concluding remarks and future extensions are given in Sec.\ref{sec:conclusion}.

\section{Review of Related Work}\label{sec:Background}
 
\subsection{Image Camouflaged Object Detection }

Camouflaged object detection (COD) has become crucial in computer vision, focusing on identifying objects hidden in an indistinguishable background. Due to the intrinsic similar patterns and unclear boundaries between objects and backgrounds, it presents a challenging task even for human eyes. A coarse-to-fine design is commonly adopted to tackle this problem. Some methods, such as SINet \cite{fan2020camouflaged}, ANet \cite{LE201945}, and PFNet \cite{9578036}, employ a joint image classification and segmentation framework. By incorporating multi-tasking analysis, these methods first identify potential regions where camouflaged objects might exist and then apply segmentation tasks to refine the initial prediction maps. DGNet \cite{ji2023gradient} decouples the task into context and texture encoders 
and links two branches by a gradient-induced transition to achieve joint feature learning. MGL \cite{9962828} and BSANet \cite{8953756} use edge details to refine segmentation decisions with boundary-aware modules. Inspired by U-Net \cite{ronneberger2015unet}, GreenCOD \cite{chen2024greencod} constructs a cascaded multi-resolution boosting network that progressively refines the prediction map as the resolution increases. 

\subsection{Video Camouflaged Object Detection}

Video Camouflaged Object Detection (VCOD) poses more challenges than image-based camouflaged object detection problems. It requires distinguishing objects in static images and dealing with dynamic scenes and potential noise introduced by the moving cameras. Combining motion cues with frame-by-frame predictions is the key to solving this problem. Conventional object detection methods explicitly calculate object motions by assessing differences between consecutive frames. Lamdouar et al. \cite{lamdouar2020betrayed} proposed a model that integrates a differentiable registration module and a motion segmentation module, utilizing optical flow and difference images obtained from the differentiable module as input to produce camouflage segmentation results. The effectiveness of this model heavily relies on the performance of the explicit motion extraction process. Instead, SLTNet \cite{cheng2022implicit} uses a dense correlation pyramid to capture motions between neighboring frames implicitly. Our approach extracts motion cues via temporal neighborhood prediction cube reconstruction without requiring extra motion estimation modules and additional computational costs.

\begin{figure*}[htbp]
\centering
{\includegraphics[width=0.8\textwidth]{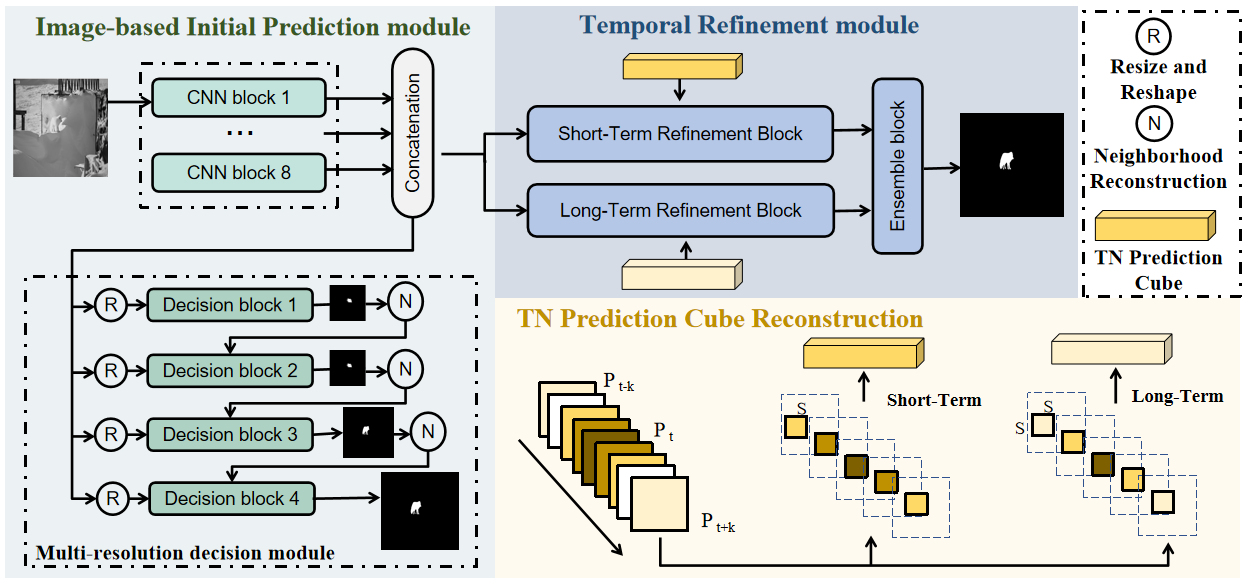}}
\caption{The system diagram of the proposed green video camouflaged object detection method (GreenVCOD).} \label{fig: diagram}
\end{figure*}

\section{GreenVCOD Method}\label{sec:method}

The green video camouflaged object detection method (GreenVCOD) is proposed in this section. First, we develop an initial-stage prediction module that extracts features and generates pixel-wise initial predictions using cascaded multi-resolution stages, as detailed in Sec.\ref{subsec:pre-trained}. Next, we incorporate temporal information to refine the initial predictions. This temporal refinement consists of the short-term and long-term refinement modules described in Sec.\ref{subsec:temporal}. In the concluding phase, a decision ensemble module, detailed in Sec.\ref{subsec:ensemble}, integrates the outcomes derived from both the short-term and long-term refinement modules, ensuring robustness and accuracy in the outcome. The overall framework is depicted in Fig.\ref{fig: diagram}.

\subsection{Initial-stage Prediction Module}\label{subsec:pre-trained}

Videos are composed of successive image sequences and most video frames contain the target camouflaged object, hence we can realize a high-quality detection on the entire video, if we achieve a high quality image-level detection. Based upon this, we first simplify the video camouflaged object detection problem into a sequence of image-based or frame-based object detection problems and develop an initial-stage prediction. The initial-stage prediction module is built based on GreenCOD \cite{chen2024greencod}. It consists of a feature extraction module and a multi-resolution decision module.

\subsubsection{Feature Extraction Module} 

The feature extraction module employs the ImageNet pre-trained EfficientNetB4 backbone to efficiently extract discriminant features for each frame. The input frames are resized into $672 \times 672 \times 3$ and fed into the EfficientNetB4 backbone for feature extraction. Next, the features from different blocks are concatenated and resized to 168 $\times$ 168 standard size. 

\subsubsection{Multi-resolution Decision Module} 

The multi-resolution decision module adopts U-Net fashion to build a multi-resolution XGBoost boosting framework. As shown in Fig.\ref{fig: diagram}, this boosting framework utilizes four cascaded classifiers, where each classifier is trained using features of a certain resolution and the prediction result from the previous classifier. The standard size features in the feature extraction module are adjusted to a certain resolution to feed into each classifier. As the cascade progresses, the resolution of the features gradually increases. Specifically, the last classifier takes features of a standard size, while the previous classifier has one-fourth the resolution of the latter. Since the predictions of each layer classifier are pixel-wise results, to enrich the prediction map fed to deeper classifiers, the predictions of the 19 $\times$ 19 neighborhood centered on that pixel are considered the prediction information passed to the next classifier. The prediction map gradually enlarges throughout the boosting process, enhancing the overall accuracy.

\subsection{Temporal Refinement Module}\label{subsec:temporal}

Video data often contains dynamic changes of the position and appearance of objects over time, hence static image information alone may not fully capture the behavior of objects. In challenging tasks like camouflaged video object detection, temporal information extraction is especially crucial. Here, we build a temporal refinement module to overcome the limitations of single-frame detection and provide more reliable and robust detection in complex scenes.

\subsubsection{Temporal Neighborhood (TN) Prediction Cube Reconstruction} 

Traditional methods incorporate individual motion detection modules, such as optical flow, to capture the motion information of successive frames. Those explicit methods consume extra computation costs, and the inaccurate motion result may lead to unreliable predictions in subsequent frames. In addition, the optical flow and the pixel-wise classifier lack necessary contextual information, while the extracted motion information cannot be directly fed into the pixel-wise classifier. To provide adequate temporal information, we have to consider the extraction of features and the estimation of the motion position, which usually requires more computing resources and running time.

\begin{figure}[htbp]
\centering
{\includegraphics[width=0.5\textwidth]{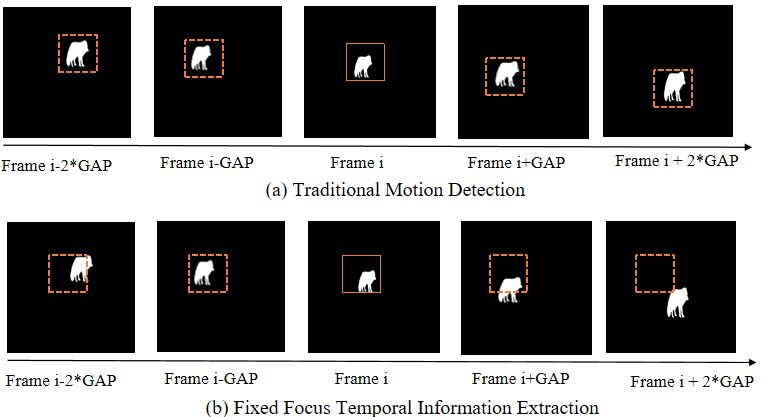}
\caption{Comparison between traditional explicit motion detection and our proposed fixed focus temporal information extraction.}} \label{fig: tcbue}
\end{figure}

To solve this problem, we propose the temporal neighborhood (TN) prediction cube to capture the temporal information implicitly. Prior works prefer to fix the background and move the focus along the movement track of the object relative to the background. Our method assumes the focus is fixed on a specific spatial location relative to the whole frame and captures the scene change within a particular time range. Since the object's movement is relatively slow between adjacent frames, fixing the focus can capture the object from different angles or focus on other parts, thereby providing adequate temporal information.

\begin{figure}[htbp]
\centering
{\includegraphics[width=0.5\textwidth]{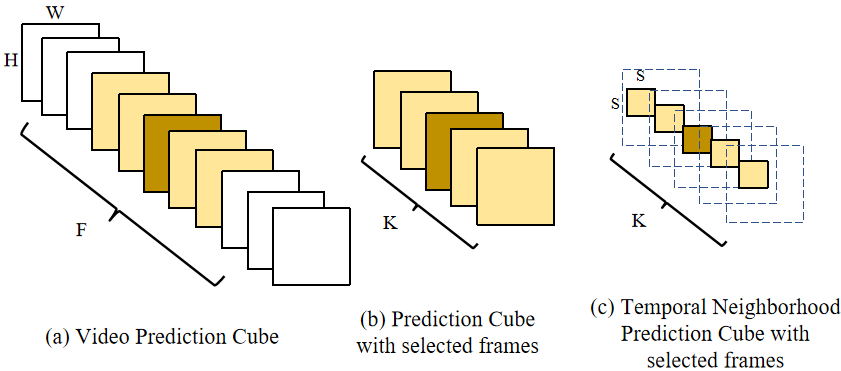}}
\caption{TN Prediction Cube Reconstruction: The brown frame represents the current frame, while the yellow frames represent the selected adjacent frames.} \label{fig: VPC}
\end{figure}

Compared with feature maps, prediction maps have only one value at each pixel location and are more efficient in providing spatial-context information. In the initial-stage XGBoost boosting framework, we adopt neighborhood predictions to enrich the feature set for decision refinement. If we consider the prediction map of each frame as an $H \times W \times 1$ cube, the spatial neighborhood is an $S \times S \times 1$ cube cropped from the original prediction cube, where the $H, W, S$ represent the image height, the image width, and the side length of the square neighborhood region. Since the depth equals one, the prediction cube only contains spatial-context information. As described in Fig.\ref{fig: VPC}, when we consider all frames in a certain video, it will construct an $H \times W \times F$ prediction cube, where $F$ is the number of frames. By extracting the neighborhood area at the same spatial location for $K$ successive frames, an $S \times S \times K$ prediction cube is constructed containing spatial-context and temporal-context information.

For each single frame, we consider the current frame, $(K-1)/2$ previous frames, and $(K-1)/2$ subsequent frames to build the TN prediction cube for the frames at the beginning and the end of the video, we will retrieve the remaining frames in reverse direction. For instance, if $K=5$ and we are at the video's final frame, we will mirror the two frames before the final frame to serve as the previous and subsequent frames of the final frame.

\subsubsection{Long-Term and Short-Term Decision Refinement}

\begin{figure}[htbp]
\centering
{\includegraphics[width=0.5\textwidth]{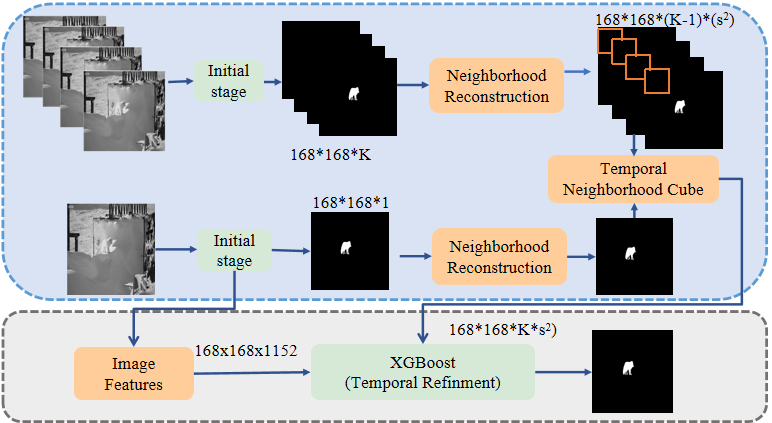}}
\caption{Temporal Refinement Module Diagram.} \label{fig: temporal}
\end{figure}

After obtaining the TN prediction cube, we reshape it into a prediction feature vector and concatenate that with the initial-stage features of the target pixel. As described in Fig.\ref{fig: temporal}, the enriched feature set trains an additional XGBoost to obtain refined decisions. By introducing motion detection or utilizing temporary information, more prosperous and more coherent contexts, and clues are captured to enable decision correction of complex frames.

Furthermore, when constructing a prediction cube, we select frames using a specific gap, referred to as $GAP$, rather than simply choosing $K$ successive frames. For instance, if we are analyzing the $i^{th}$ frame with $GAP=2$, the frames used to construct the prediction cube would be the $i-4^{th}, i-2^{th}, i^{th}, i+2^{th}, i+4^{th}$ frames. The conventional strategy of considering successive $K$ frames is a case where $GAP=1$. Building upon this, we refer to this approach as short-term when $GAP=1$, and we call it long-term when $GAP > 1$. Since the long-term and short-term refinement modules only require different parameterized construction in the temporal prediction cubes, extracting temporal information under different term settings will not cause additional computation.

\subsection{Decision Ensemble Module}\label{subsec:ensemble}

The time range required to calculate adequate temporal information varies significantly in video object detection. Fast-moving targets necessitate a shorter time range to effectively capture meaningful information, while slow-moving targets benefit from a longer detection time. Our approach integrates short-term and long-term refinement strategies, where both refinement modules are built in a parallel architecture. Furthermore, we enhance our method by integrating prediction maps derived from these refinement modules and employing adaptive thresholds to balance the tradeoffs between short-term and long-term decisions, ensuring optimal performance across varying motion patterns.

\section{Experiments}\label{sec:experiment}

We demonstrate the performance evaluation of our proposed framework on
by MoCA-Mask dataset \cite{cheng2022implicit} in this section.  

\subsection{Datasets}

\textbf{MoCA-MASK} is generated based on the original Moving Camouflaged Animals (MoCA) Dataset. The original ground truth bounding box is extended to dense segmentation masks for more comprehensive performance evaluation. The training set includes 71 videos with 19,313 frames, and the testing set contains 16 video sequences with 3,626 frames. 

\textbf{COD10K} is one of the largest camouflage object detection datasets. It consists of 5066 camouflaged images and 1934 non-camouflaged images. The frame-based models within our initial-stage prediction modules are pre-trained on the COD10K \cite{fan2020camouflaged} dataset.

\subsection{Experimental Setup}

We conduct the experiments on MoCA-Mask dataset. Since this dataset has insufficient frame and scene diversity and small camouflage objects, we choose to pre-train the cascaded multi-resolution decision part in the initial-stage prediction module is pre-trained on COD10K and CMAO \cite{LE201945} dataset. At the same time, the temporal refinement module is trained and fine-tuned only with MoCA-Mask training data. The final prediction map of each frame is of standard size 168 $\times$ 168, which will be resized back to its original frame size for performance evaluation. For a fair comparison with state-of-the-art, we utilize five evaluation metrics:  enhanced-alignment measure ($E_\phi$), weighted F-measure ($F^\beta_w$), mean absolute error (MAE), mean Dice (mDice) and the mean intersection of union (mIoU).

\subsection{Performance Evaluation}

\subsubsection{Comparisons with State-of-The-Arts}

\begin{figure}[htbp]
\centering
{\includegraphics[width=0.48\textwidth]{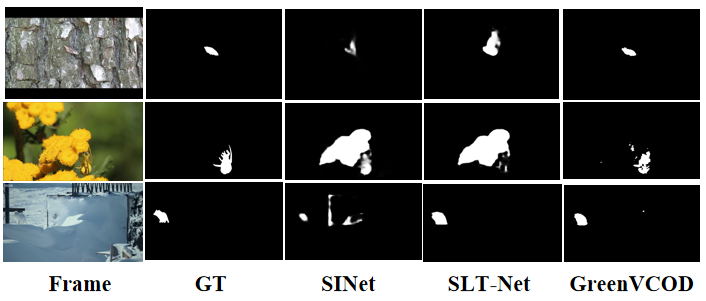}}
\caption{Visual comparisons on example video frames with state-of-the-art camouflage detection methods.} \label{fig: v}
\end{figure}

As shown in Table \ref{tab:performance}, we compare our proposed method with 6 state-of-the-art methods under MoCA-Mask dataset \cite{cheng2022implicit}. Our proposed method outperforms all methods on MAE (0.0079) and $F^{\beta}_w$ (0.311), while we rank second in mean dice and average IOU. Notably, our approach is comparable in performance to recently proposed implicit motion-extraction-based methods, such as SLT-Net \cite{cheng2022implicit}. Fig.\ref{fig: v} provides several visual examples to assess prediction quality. Compared to the most advanced detection methods, our method has more accurate detection results and higher visual quality, especially in the details of object shapes and more precise detection boundaries.

\begin{table}[htbp]
\caption{Quantitative results on the MoCA-Mask Dataset. The best-performing method of each
category is highlighted in bold, while the second performing method is highlighted with underlines.} 
\label{tab:performance}
\begin{center}
\begin{tabular}{c|c|c|c|c|c}\hline
& \multicolumn{5}{c} {MoCA-Mask w/o pseudo labels \#}\\ \cline{2-6} 
& $F^{\beta}_w \uparrow$ & $E_\phi \uparrow$ & $MAE\downarrow$ & $mDice\uparrow$ & $mIoU\uparrow$\\ \hline
SINet \cite{fan2020camouflaged} & 0.231 & \underline{0.699} & 0.028 & 0.276 & 0.202 \\ 
SINet-v2 \cite{fan2021concealed} &  0.204 & 0.642 & 0.031 & 0.245 & 0.180 \\ \hline
PNS-Net \cite{ji2021progressively} & 0.097 & 0.510 & 0.033 & 0.121 & 0.101 \\ 
RCRNet \cite{yan2019semi} & 0.138 &  0.527 &  0.033 & 0.171 & 0.116 \\
MG \cite{9711323} & 0.168 &  0.561&  0.067 & 0.181 & 0.127\\ 
SLT-Net \cite{cheng2022implicit} & \textbf{0.311} & \textbf{0.759} & \underline{0.027} & \textbf{0.360} & \textbf{0.272} \\ \hline
Ours & \textbf{0.311} & 0.635 & \textbf{0.00786} & \underline{0.318} & \underline{0.243}\\\hline
\end{tabular}
\end{center}
\end{table}

\begin{figure*}[htbp]
\centering
{\includegraphics[width=0.75\textwidth]{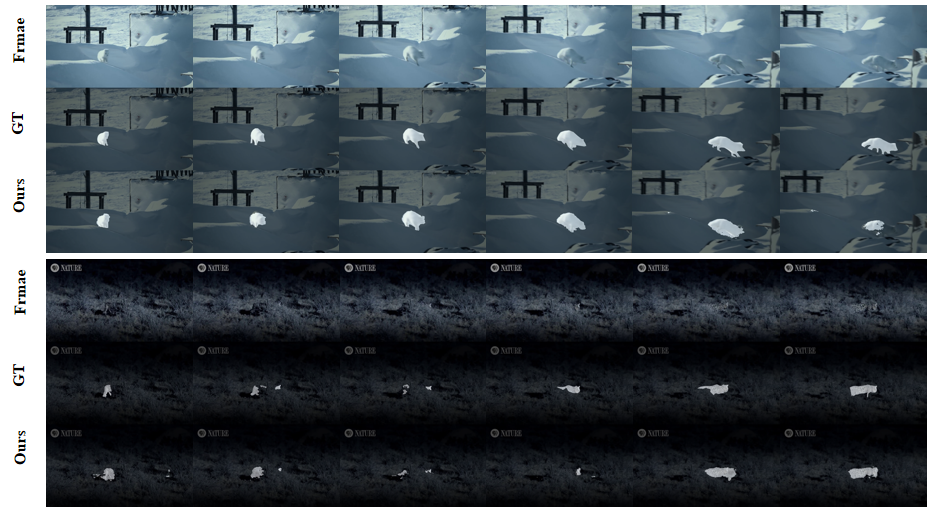}}
\caption{Visual comparisons on consecutive video frames.} \label{fig: frames}
\end{figure*}

We also show some sample video frames in Fig.\ref{fig: frames} to demonstrate performance on continuous video frames. For a more comprehensive evaluation, we provide two sample videos with different difficulty levels and focus, the Arctic Fox and Sand Cat videos. In the first video, the Arctic fox's shape and shooting angle change during rapid movement, while in the second video, the shading of bushes and intersecting light conditions make even ground truth unable to provide a clear outline of the sand cat. The experimental results demonstrate that our model can detect the camouflaged object stably and get clear moving animal boundaries under different challenging modes.

\subsubsection{Ablation Study: Short-term and Long-term Modules} 

For decision refinement, we use short - and long-term modules with parallel structures to balance our attention to the speed of motion of different objects. As shown in Table \ref{tab:ablation}, long-term optimization results outperform short-term optimization results in almost all evaluations. Because the different temporal modules extract complementary context information, the final ensemble result from the fusion of both optimizations is better than any single one in all evaluations.

\begin{table}[htbp]
\caption{Ablation studies of short-term and long-term architectures on MoCA-Mask test set.} 
\label{tab:ablation}
\begin{center}
\begin{tabular}{c|c|c|c|c|c}\hline
& \multicolumn{5}{c} {MoCA-Mask w/o pseudo labels \#}\\ \cline{2-6} 
& $F^{\beta}_w \uparrow$ & $E_\phi \uparrow$ & $MAE\downarrow$ & $mDice\uparrow$ & $mIoU\uparrow$\\ \hline
Ours (Short-Term) & 0.284 & 0.630 & 0.00828 & 0.289 & 0.216\\
Ours (Long-Term) & 0.304 & 0.629 & 0.00763 & 0.314 & 0.240\\
Ours (Ensemble) & 0.311 & 0.635 & 0.00786 & 0.318 & 0.243\\\hline
\end{tabular}
\end{center}
\end{table}

\subsubsection{Computational Complexity in Inference}
We calculate the model size and computational complexity of the proposed GreenVCOD framework, where the complexity is quantified through Multiply-Accumulate Operations (MACs) for evaluation. As shown in the table. \ref{tab:model}, the model size of GreenVCOD is approximately 19.9M, which is comparable to the smaller-sized image detection method SINet-v2 (26.98M). In addition, our model requires 17.42G MACs in the inference phase, of which 77.5\% of the computational cost is due to the EfficientNet backbone. In contrast, the TN prediction cube is generated based on a prediction graph of successive frames and does not generate additional calculations, making the decision refinement module highly efficient.

\begin{table}[htbp]
\caption{The number of parameters and computational complexity of GreenVCOD.} 
\label{tab:model}
\begin{center}
\begin{tabular}{c|c|c|c|c}

\hline
Module & Sub-Module & $\#$ Trees & Depth &  Parameters (\%) \\ \hline
Stage-1 & EfficientNetB4 & \_ & \_ & 16,742,216 (84.1\%) \\
Stage-1 & XGBoost1  & 10000 & 3 & 220,000 (1.1\%) \\
Stage-1 & XGBoost2  & 10000 & 3 & 220,000 (1.1\%)\\ 
Stage-1 & XGBoost3  & 10000 & 3 & 220,000 (1.1\%)\\ 
Stage-1 & XGBoost4  & 10000 & 3 & 220,000 (1.1\%)\\ \hline
Stage-2 & XGBoost(long-term)  & 6000 & 6 & 1,140,000 (5.7\%) \\
Stage-2 & XGBoost(short-term)  & 6000 & 6 & 1,140,000 (5.7\%)\\ \hline
Total & \_ & \_  & \_ & 19,902,216 \\ \hline
\end{tabular}

\vspace{0.5cm}
\begin{tabular}{c|c|c|c}
\hline
Module & Sub-Module & Map size & MACs (\%)\\ \hline
Stage-1 & EfficientNetB4 & \_ & 13,503,446,880 (77.5\%) \\
Stage-1 & XGBoost1  & $42\times42$ & 70,560,000 (0.4\%) \\
Stage-1 & XGBoost2  & $42\times42$  & 70,560,000 (0.4\%)\\ 
Stage-1 & XGBoost3  & $84\times84$ & 282,240,000 (1.6\%)\\ 
Stage-1 & XGBoost4  & $168\times168$ & 1,128,960,000 (6.4\%)\\ \hline
Stage-2 & XGBoost(long-term)  & $168\times168$ & 1,185,408,000 (6.8\%) \\
Stage-2 & XGBoost(short-term)  & $168\times168$ & 1,185,408,000 (6.8\%)\\ \hline
Total & \_ & \_ & 17,426,582,880 \\ \hline

\end{tabular}
\end{center}
\end{table}


\section{Conclusion and Future Work}\label{sec:conclusion}

GreenVCOD was proposed for camouflage object detection in video footage in this work. It utilized an image-based COD method to obtain the initial prediction map and optimized predictions with short- and long-term ensembles. GreenVCOD offers competitive predictive performance and impressive efficiency compared to state-of-the-art VCOD methods. Due to the limitations of the MoCA-Mask dataset, most existing methods often exploited partial pre-training with more diverse datasets. As future extensions, we will address the few-shot camouflage detection problem.



\end{document}